\documentclass{article}

    \PassOptionsToPackage{numbers, compress}{natbib}
 \usepackage[dblblindworkshop, final]{neurips_2025}
\workshoptitle{The Second Workshop on GenAI for Health: Potential, Trust, and Policy Compliance}

\usepackage[utf8]{inputenc} % allow utf-8 input
\usepackage[T1]{fontenc}    % use 8-bit T1 fonts
\usepackage{hyperref}       % hyperlinks
\usepackage{url}            % simple URL typesetting
\usepackage{booktabs}       % professional-quality tables
\usepackage{amsfonts}       % blackboard math symbols
\usepackage{nicefrac}       % compact symbols for 1/2, etc.
\usepackage{microtype}      % microtypography
\usepackage{xcolor}         % colors
\usepackage{amsmath}
\usepackage{graphicx}
\usepackage{caption}
\usepackage{multirow}

\title{Bridging Graph and State-Space Modeling for Intensive Care Unit Length of Stay Prediction}

\author{%
  Shuqi Zi\textsuperscript{1} \quad
  Haitz S\'aez de Oc\'ariz Borde\textsuperscript{2} \quad
  Emma Rocheteau\textsuperscript{1} \quad
  Pietro Li\`o\textsuperscript{1} \\
  \\
  \textsuperscript{1} University of Cambridge, UK \\
  \textsuperscript{2} University of Oxford, UK \\
}

\begin{document}

\maketitle

\begin{abstract}
Predicting a patient’s length of stay (LOS) in the intensive care unit (ICU) is a critical task for hospital resource management, yet remains challenging due to the heterogeneous and irregularly sampled nature of electronic health records~(EHRs). In this work, we propose S$^2$G-Net, a novel neural architecture that unifies state-space sequence modeling with multi-view Graph Neural Networks~(GNNs) for ICU LOS prediction. The temporal path employs Mamba state-space models~(SSMs) to capture patient trajectories, while the graph path leverages an optimized GraphGPS backbone, designed to integrate heterogeneous patient similarity graphs derived from diagnostic, administrative, and semantic features. Experiments on the large-scale MIMIC-IV cohort dataset show that S$^2$G-Net consistently outperforms sequence models (BiLSTM, Mamba, Transformer), graph models (classic GNNs, GraphGPS), and hybrid approaches across all primary metrics. Extensive ablation studies and interpretability analyzes highlight the complementary contributions of each component of our architecture and underscore the importance of principled graph construction. These results demonstrate that S$^2$G-Net provides an effective and scalable solution for ICU LOS prediction with multi-modal clinical data. The code can be found at~\url{https://github.com/ShuqiZi1/S2G-Net}.
\end{abstract}

\section{Introduction}

There has been a growing focus on predicting patient outcomes in intensive care units (ICUs), along with increasing research interest in applying Graph Deep Learning techniques for this purpose~\cite{rocheteau2021predicting}. This surge in attention is largely driven by the expanded use of Electronic Health Records (EHRs), the demonstrated effectiveness of graph representation learning across diverse domains—including recommendation systems and misinformation detection~\cite{monti2019fakenewsdetectionsocial}, financial forecasting~\cite{zhang2023graphneuralnetworksforecasting}, and gene regulatory network modeling~\cite{borde2025neural}, to name a few—and the broader effort to reduce preventable deaths through early warning systems~\cite{Moon2011AnEY}.

In this work, we bridge recent techniques in the Graph Neural Network (GNN) literature with sequence modeling methods by using graphs to model inter-patient correlations and sequence models to capture time-series patient data, enabling accurate predictions of ICU length of stay (LOS) for improved hospital resource allocation, management, and optimization. Despite recent progress, which we later discuss in Section~\ref{Deep Learning Approaches for Clinical Data}, several challenges remain. Firstly, there is a clear gap in the literature for general-purpose models that jointly capture both temporal patient trajectories and inter-patient correlations, particularly for tasks like ICU LOS prediction, where existing approaches tend to focus on either temporal or spatial aspects in isolation or are tailored to narrow clinical outcomes such as mortality or disease progression. Secondly, traditional GNNs generally operate on single-view, static graphs, which are ill-suited for modeling heterogeneous, multi-modal clinical features and dynamically evolving patient relationships. Transformer-based graph backbones (e.g., GraphGPS~\cite{rampavsek2022recipe} and other Graph Transformers~\cite{elucidating}) are powerful but computationally demanding, with quadratic complexity due to all-to-all node-wise communication, and they often fail to accommodate the unique characteristics of ICU data, such as sparsity and multi-view structure. Moreover, many existing models lack sufficient interpretability~\cite{Molnar2020InterpretableML} (which is also critical in many real-world ML applications such as engineering~\cite{Borde01022022} or finance~\cite{brigo2021interpretabilitydeeplearningfinance}), limiting their clinical relevance.

In this study, we propose S$^2$G-Net (State-space × Optimized Multi-View GNN) for ICU LOS prediction. It comprises three components: a temporal pathway using a state-space backbone (Mamba \cite{Gu23}) to handle irregular and long-range temporal dependencies in patient time-series data; a graph pathway employing an optimized GraphGPS to learn from multiple patient similarity graphs built from diagnostic, semantic, and administrative views; and an auxiliary static-feature branch.

The main contributions of this work are:

\begin{itemize}
    \item We propose a method for constructing heterogeneous patient similarity graphs using diagnostic, semantic, and administrative features, enabling richer representations of clinical relationships compared to single-view or static graphs.
    \item We develop a dual-path neural architecture that unifies temporal dynamics with population-level similarity modeling, tailored to the heterogeneous structure of ICU data.
    \item We adapt and enhance the GraphGPS architecture for clinical use, replacing Transformer~\cite{vaswani2017attention} layers with state-space models (SSMs) and introducing more efficient graph fusion strategies.
    \item We benchmark S$^2$G-Net against strong sequence, graph, and hybrid baselines on the large-scale MIMIC-IV dataset~\cite{johnson2023mimic}, demonstrating superior performance in terms of accuracy, resource efficiency, and calibration.
    \item We conduct feature attribution, ablation, and calibration analyzes to evaluate the interpretability, reliability, and robustness of the model under real-world conditions.
\end{itemize}

\section{Preliminaries and Related Work}
\label{relatedwork}

\subsection{Understanding the Complexities of Length of Stay Prediction}

An ICU is a specialized healthcare unit providing advanced monitoring, physiologic organ support, and expert medical and nursing care for critically ill patients~\cite{marshall2017intensive}. Globally, ICU resources are limited and expensive. Bed capacity varies by region, from 4.2 beds per 100,000 people in Portugal to 29.2 in Germany~\cite{rhodes2012variability}, with pandemics like COVID-19 pushing occupancy above 95\% in some regions~\cite{grasselli2020critical}. The daily cost per ICU patient cared for in developed countries ranges from \$2,000 to \$5,000 per day~\cite{chacko2023approach, guest2020modelling, carrandi2024costs, kilicc2019cost}, while prolonged hospital stays significantly increase the financial burden on both hospitals and families. In the United States itself, ICU costs contribute nearly 20 per cent of hospital costs~\cite{wagner1983hidden}, amounting to over \$100 billion annually~\cite{kannan2023growth}. Beyond the costs, ICU LOS is closely tied to patient outcomes. Prolonged LOS is related to complications like hospital acquired infections~\cite{toh2017factors}, delirium, and muscle atrophy, while early discharge can lead to readmission or worsening of the prognosis~\cite{loreto2020early}. Research has shown that in developed healthcare systems, ICU stays typically range from 3 to 8 days~\cite{moitra2016relationship}, but vary widely depending on the condition. For example, patients recovering from cardiac surgery may require a 1-2 day hospital stay, whereas patients with sepsis or acute respiratory distress syndrome~(ARDS) may require weeks of intensive care. Accurate LOS prediction can help reduce inefficiencies, improve patient care, and alleviate the financial and operational burden on healthcare systems.

In fact, the dynamic and heterogeneous nature of ICU data makes predicting ICU LOS inherently challenging. ICU patients are continuously monitored, generating high-frequency time-series data including vital signs (e.g., heart rate, blood pressure)~\cite{MonitoringTestingCritical}, laboratory results (e.g., blood gases, inflammatory markers), and interventions (e.g., medications, mechanical ventilation). These data display complex and intricate dependencies over time and across organ systems. For example, oxygenation may improve with diuretics in patients with heart failure, whereas worsening renal function may coincide with hypotension or sepsis progression, creating a highly non-linear and time-dependent problem. Moreover, in real ICU settings, only partial data points are recorded at uniform intervals, with physiological parameters like continuously monitored heart rate and oxygen saturation providing high-frequency data, while laboratory test results like blood gas levels or electrolyte measurements are recorded intermittently. From a deep learning perspective, these characteristics present several challenges: ICU records are often irregular and incomplete, with certain variables sampled at widely varying frequencies. The data are also highly heterogeneous, varying across patients, conditions, and treatments. This makes it difficult for models to learn consistent patterns. Moreover, the relationships between variables are often non-linear and evolve over time, requiring models to capture both short- and long-term dependencies. These challenges are further complicated by the need for models to be accurate, interpretable, and robust in real-world clinical settings.

\subsection{Deep Learning Approaches for Clinical Data}
\label{Deep Learning Approaches for Clinical Data}

\paragraph{Temporal Sequence Models.}Recurrent neural network~(RNN)~\cite{rumelhart1986learning}, particularly Long Short-Term Memory~(LSTM)~\cite{Hochreiter97} models, are widely used for patient time-series modeling~\cite{liu2022multi, ma2017dipole, rajkomar2018scalable}. For example, Harutyunyan et al.~\cite{Harutyunyan19} provided LSTM baselines for mortality, decompensation, length of stay, and phenotype prediction on MIMIC-III~\cite{johnson2016mimic}. LSTMs can model sequential dependencies and have been adapted for multivariate and irregular EHR data using masking and imputation~\cite{Lipton16}. However, their performance declines with long and highly irregular ICU sequences. Recently, transformer architectures have shown promise in EHR modeling by capturing long-range dependencies through attention mechanisms. Models such as BEHRT~\cite{Li20} and Med-BERT~\cite{Rasmy21} adapt transformers to structured clinical data, while recent approaches introduce positional encodings and sparse attention for handling irregular sampling~\cite{Song18}. Nonetheless, transformers are computationally intensive due to quadratic self-attention and can overfit with limited clinical data. Structured SSMs, such as S4 and Mamba, complement RNNs and transformers. S4~\cite{Gu22S4} achieves linear-time sequence modeling and maintains long-term dependencies using structured kernels. Mamba~\cite{Gu23} introduces input-dependent recurrence for greater stability and efficiency on long sequences. Both have been applied to domains such as genomics and time-series analysis, but applications in high error rate ICU data remain rare.

\paragraph{Graph-Based Patient Modeling.}Patient similarity graphs can model complex multi-modal relationships between individuals. Parisot et al.~\cite{Parisot18} applied graph convolutional networks~(GCNs) to patient graphs for disease classification, constructing edges from demographic and imaging features. Subsequent studies have extended graph-based methods~\cite{kipf2016semi,borde2023projections} to tasks such as readmission prediction~\cite{Golmaei21}, medication recommendation~\cite{shang2019gamenet}, sepsis risk forecasting~\cite{Yin25}, as well as Alzheimer's disease and mild cognitive impairment prediction~\cite{borde2023latent}, demonstrating improvements over models that treat samples independently. Recent advances include dynamic and heterogeneous GNNs that capture temporal and attribute changes within patient cohorts. Dynamic GNNs model evolving edges and features over time~\cite{zheng2025survey}, whereas multi-view GNNs integrate diverse data modalities (e.g., ICD codes, labs, text) for richer representations~\cite{kim2023heterogeneous, Liu20heter}. Nevertheless, most clinical applications to date still use static or single-view graphs due to computational and modeling complexity. 
Beyond these considerations, another challenge lies in balancing local neighborhood aggregation and global context modeling. Local-Global GNNs such as GraphGPS~\cite{rampavsek2022recipe} and GLGNN~\cite{Eliasof24} address this issue by combining message passing with global attention modules.
These approaches are well-suited to clinical graphs, where both local similarity and global cohort structure are relevant for prediction.

% \paragraph{Other Clinical Prediction Tasks.} RNNs, LSTM, and Temporal Convolutional Networks (TCNs)~\cite{lea2016temporal} have shown promise in modeling time-series data for ICU outcome prediction. Mili et al.\cite{mili2022icu} proposed an LSTM-based approach to predict length of stay (LOS) and in-hospital mortality using the MIMIC-III dataset~\cite{johnson2016mimic}. Graph-based methods have also gained traction in this domain. The works of Zafeiropoulos~\cite{Scarselli09} and Sun et al.\cite{sun2020disease} applied Graph Neural Networks (GNNs)~\cite{zhou2020graph} to static medical data for tasks such as disease surveillance and prediction. Similarly, latent graph inference methods have been used to predict Alzheimer’s disease and mild cognitive impairment~\cite{Kazi_2023,borde2023latent,borde2023projections}. Temporal GNNs that account for time-evolving features or longitudinal imaging data have also been explored. For instance, Brain Latent Progression (BrLP)~\cite{puglisi2024enhancing} predicts disease evolution at the individual level using 3D brain MRI, incorporating a priori knowledge to improve accuracy. Likewise, Wang et al.\cite{wang2023spatio} utilize spatio-temporal similarity measurements of brain biomarkers to forecast the progression of Alzheimer’s disease.

\paragraph{Hybrid Temporal-Graph Models.}Models that jointly learn from temporal patient trajectories and inter-patient graph structure have shown improved performance in clinical forecasting tasks. GAMENet~\cite{shang2019gamenet} integrates GCNs with dynamic memory modules for medication recommendation, and LSTM-GNN architectures~\cite{rocheteau2021predicting} combine sequence modeling with graph-based neighborhood aggregation for ICU outcome prediction. Despite progress, most existing models inadequately integrate temporal and relational information, lack multi-modal graph construction, or are computationally inefficient. We address these gaps with S$^2$G-Net, a unified and efficient hybrid framework.

\section{Methods}

\subsection{Problem Definition}

We consider the task of predicting the ICU LOS for a set of patients using multi-modal clinical data collected during the first 48 hours of admission. Let $\mathcal{G} = (\mathcal{V}, \mathcal{E})$ be a patient similarity graph, where each node $v_i \in \mathcal{V}$ represents a unique ICU stay, and $(v_i, v_j) \in \mathcal{E}$ indicates clinical similarity based on diagnosis codes, semantic embeddings, or administrative metadata. 
Each node $v_i$ is associated with: (i) A multivariate time series $\mathbf{X}_i^{\mathrm{TS}} \in \mathbb{R}^{T \times d}$, where $T = 48$ is the number of hourly time steps and $d$ is the number of variables (labs, vitals); (ii) A static feature vector $\mathbf{x}_i^{\mathrm{Flat}} \in \mathbb{R}^{d_f}$, formed by concatenating demographics, admission metadata, and diagnosis indicators; (iii) A continuous label $y_i \in \mathbb{R}_+$, representing the true ICU LOS in days. 
The objective is to learn a function $F_\Theta$ that maps patient data and graph structure to a LOS prediction: $\hat{y}_i = F_\Theta(\mathbf{X}_i^{\mathrm{TS}}, \mathbf{x}_i^{\mathrm{Flat}}, A(\mathcal{G}))$, where $A(\mathcal{G})$ is the adjacency matrix of the graph. Model parameters $\Theta$ are optimized to minimize the prediction error over all patients: $\min_{\Theta} \; \frac{1}{N} \sum_{i=1}^{N} \mathcal{L}(\hat{y}_i, y_i),$ where $\mathcal{L}(\cdot, \cdot)$ is a regression loss. All features are extracted strictly within the first 48 hours to ensure prospective prediction and avoid information leakage. LOS targets are transformed using $\log(1 + y_i)$ to facilitate stable learning, and predictions are reverted with $\exp(\hat{y}_i) - 1$.

\subsection{Patient Graph Construction} \label{sec:graph}
\begin{figure}[htbp]
    \centering
    \includegraphics[width=\textwidth, trim = {2cm 12cm 1.7cm 1.65cm}, clip]{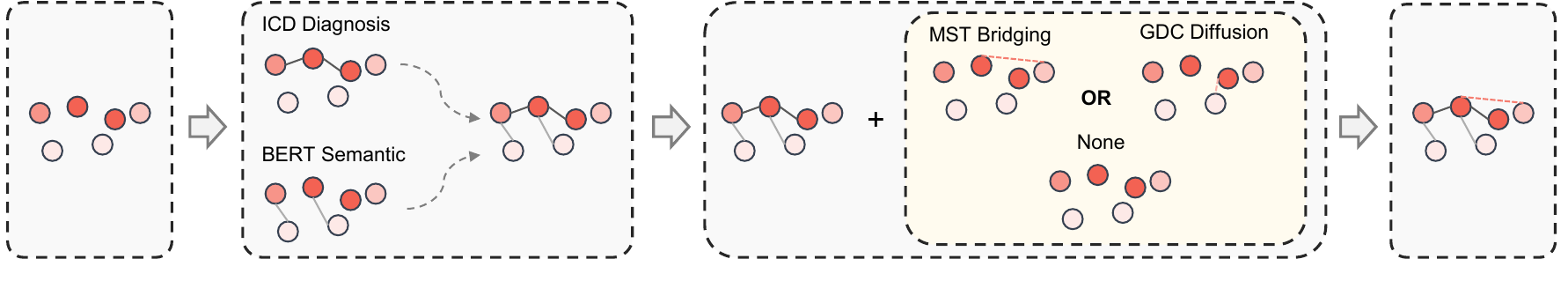}
    \caption{
    Multi-view graph construction and augmentation pipeline. Graphs based on ICD codes and BERT semantics are merged to form an initial patient graph. Connectivity and global structure are optionally enhanced via MST bridging or GDC. The final graph captures local, semantic, and long-range relations for downstream learning.
    }
    \label{fig:fusion_pipeline}
\end{figure}

To capture structured inter-patient relationships in ICU data, we construct a multi-view patient similarity graph $\mathcal{G}$ that integrates both structured diagnostic information and semantic similarity. Figure~\ref{fig:fusion_pipeline} illustrates the fusion process. The adjacency matrix for this graph serves as input to the model graph encoder and provides inductive relational bias by embedding both clinical and linguistic proximity among patients. The final graph defines four edge types: Type-0 for diagnosis-based similarity, Type-1 for BERT-based semantic similarity, and two optional augmentation types, Type-2 for minimum spanning tree (MST) and Type-3 for graph diffusion convolution (GDC). 

\paragraph{Diagnosis-Based Similarity Graph.}We encode structured clinical similarity by constructing a graph from filtered ICD-9/10 codes. Each patient is represented by a sparse binary vector $\mathbf{d}_s \in \{0,1\}^{d_d}$, indicating assigned specific diagnosis. Edges are constructed based on pairwise similarity using one of three strategies: (i) TF-IDF Cosine Similarity, where diagnosis vectors are reweighted by inverse document frequency and cosine similarity selects the top-$k$ neighbors; (ii) FAISS Approximate KNN \cite{johnson2019billion}, which performs efficient inner product search over large patient sets; and (iii) Penalized Co-occurrence, which favors patients with high diagnostic overlap while penalizing multi-morbidity noise. After similarity calculation, we prune each node to its top-$k_{\text{diag}}$ neighbors and normalize edge weights via $\log(1+x)$ or z-score. We remove the bottom 30\% of weak connections to improve sparsity. The resulting graph is denoted $\mathcal{G}_{\text{diag}} = (\mathcal{V}, \mathcal{E}_{\text{diag}})$.

\paragraph{BERT Semantic Graph.}

To capture semantic similarity beyond diagnosis code overlap, we construct a graph using sentence embeddings derived from patients’ diagnosis descriptions. For each patient, diagnosis codes are concatenated into a single text sequence, truncated or zero-padded to 512 tokens. A pre-trained DistilBERT model \cite{sanh2019distilbert,devlin2019bert} encodes each input into a fixed-length embedding $\mathbf{b}_s \in \mathbb{R}^{768}$. Patient similarity is computed via Gaussian-kernelized Euclidean distance:

\begin{equation}
w_{ij}^{\text{bert}} = \exp\left( -\frac{||\mathbf{b}_i - \mathbf{b}_j||_2^2}{2\sigma^2} \right)
\end{equation}

where $\sigma$ is set to the median pairwise distance:
$\sigma = \text{median}_{i \neq j}(|\mathbf{b}_i - \mathbf{b}_j|_2)/\sqrt{2}$, 
following \cite{zelnik2004self}. Top-$k_{\text{bert}}$ neighbors are retained, and the resulting graph is $\mathcal{G}_{\text{bert}} = (\mathcal{V}, \mathcal{E}_{\text{bert}})$.

\paragraph{Multi-View Graph Fusion.}

The initial graph $\mathcal{G} = (\mathcal{V}, \mathcal{E})$ is formed by merging $\mathcal{E}_{\text{diag}}$ and $\mathcal{E}_{\text{bert}}$, retaining edge types for modality-aware message passing. To ensure global connectivity and capture long-range dependencies~\cite{liang2025quantifyinglongrangeinteractionsgraph}, we optionally apply two rewiring strategies: MST-based bridging to link disconnected components, and GDC-based diffusion using first-order Personalized PageRank \cite{gasteiger2019diffusion}. 
A maximum out-degree of 15 is enforced via stratified edge-type sampling to preserve diversity.

% To solve potential graph fragmentation, we implement two optional rewiring techniques: 
% (1) Minimum Spanning Tree (MST). To prevent fragmentation, we connect disconnected components via low-cost edges based on inverted similarity. Added edges are marked as Type-2.
% (2) Graph Diffusion Convolution (GDC). We apply first-order Personalized PageRank \cite{gasteiger2019diffusion}:$   A_{\text{diffused}} = \alpha I + (1-\alpha) A_{\text{normalized}}, \quad \alpha = 0.05,$ to propagate local similarity signals globally. Added edges are marked as Type-3. 

\subsection{Model Architecture}

\begin{figure*}[htbp]
    \centering
    \includegraphics[width=\textwidth,trim = {2cm 0cm 3cm 0cm}, clip]{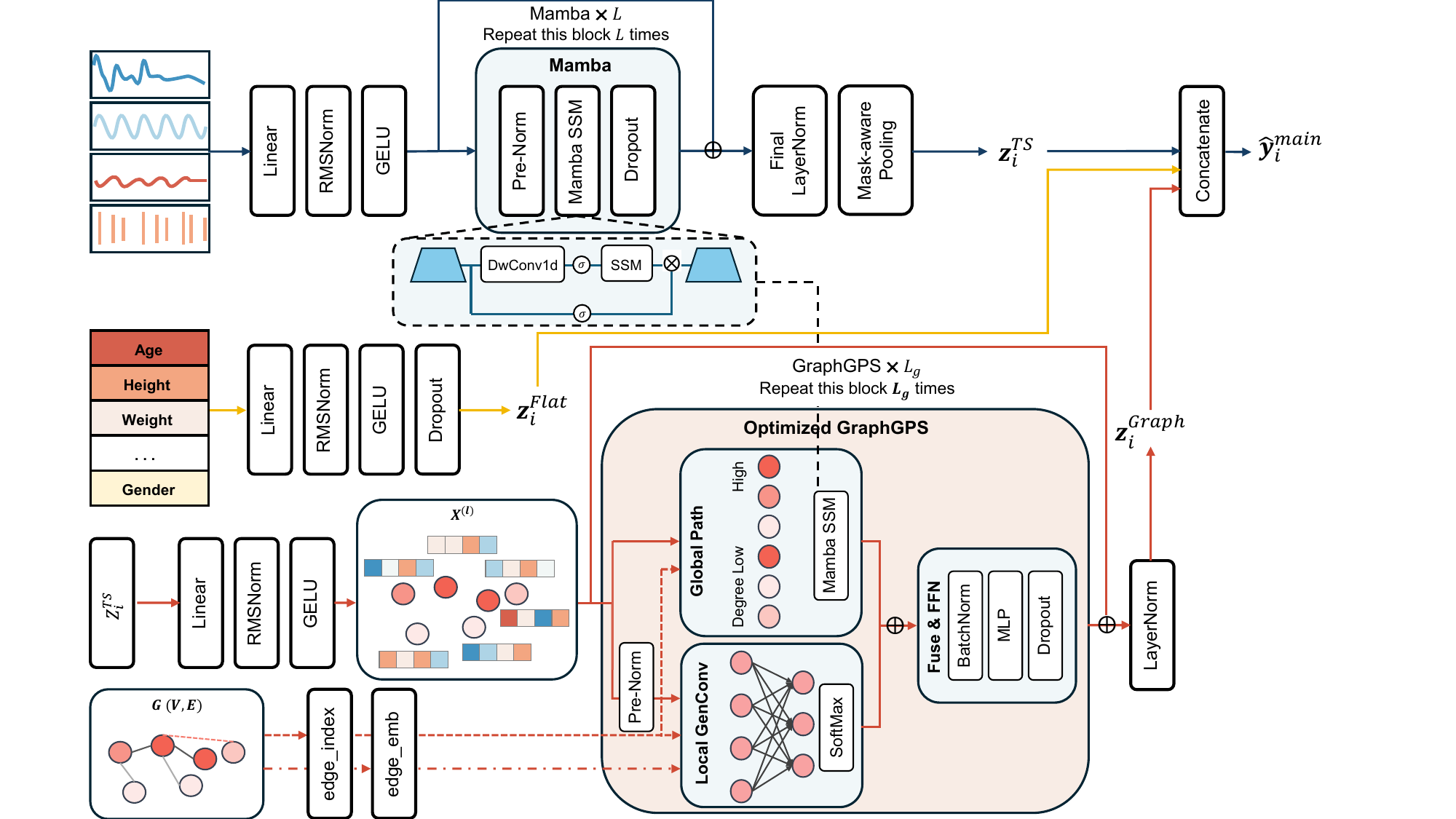}
        \caption{
            S$^2$G-Net model architecture. It consists of three branches: (i) a temporal encoder based on SSMs, (ii) a multi-view graph encoder capturing patient similarities, and (iii) a flat feature encoder for static attributes. The final representation is formed via weighted fusion of these components.
        }
    \label{fig:model_arch}
\end{figure*}

As illustrated in Figure~\ref{fig:model_arch}, S$^2$G-Net is a dual-path neural architecture consisting of temporal encoders and multi-view graph encoders, augmented with flat feature encoders for static attributes.

\paragraph{Temporal Encoder (State-Space).}
Time-series data in EHRs is typically long, sparse, and noisy. To capture long-range dependencies efficiently, we adopt the Mamba state-space model as the temporal encoder. Given a patient $i$ with multivariate time-series $\mathbf{X}_i^{\mathrm{TS}} \in \mathbb{R}^{T \times d}$ over the first 48 hours, we first apply a linear projection and normalization:
\begin{equation}
    \mathbf{H}_i^{(0)} = \mathrm{GELU}\left(\mathrm{RMSNorm}\left(\mathbf{X}_i^{\mathrm{TS}} \mathbf{W}_0 + \mathbf{b}_0\right)\right).
\end{equation}

This sequence is then processed through $L$ stacked Mamba blocks to model temporal patterns. To produce a fixed-length representation while handling missing values and variable lengths, we apply a mask-aware pooling function over the temporal dimension:

\begin{equation}
\mathbf{z}_i^{\mathrm{TS}} = \mathrm{MaskPool}\left( \mathbf{H}_i^{(L)}, \mathbf{m}_i \right)
\end{equation}
where $\mathbf{m}_i \in \{0,1\}^{T}$ is a binary mask indicating observed time steps.

% To handle missing data and variable sequence lengths, we apply a mask-aware pooling operation over the temporal axis to obtain a fixed-length embedding $\mathbf{z}_i^{\mathrm{TS}}$.

\paragraph{Graph Encoder (Local GENConv + Global Mamba).}
To capture population-level relational structure, we employ the multi-view patient similarity graph $\mathcal{G} = (\mathcal{V}, \mathcal{E})$ introduced in Section~\ref{sec:graph}. Nodes represent individual patients, and edges encode heterogeneous similarity relations with typed and weighted attributes. Each node $i$ is initialized with its temporal embedding $\mathbf{z}_i^{\mathrm{TS}}$, then projected via a two-layer MLP with LayerNorm to produce $\mathbf{x}_i^{(0)}$.

We stack $L_g$ GraphGPS blocks. In each block $\ell\in\{0,\dots,L_g-1\}$, we compute a local update with GENConv layers that incorporate typed and weighted edge attributes, and a global update with a Mamba state-space encoder applied to a degree-ordered node sequence with slight random perturbations to resolve ties. We denote the local and global features by $\mathbf{h}_i^{(\ell)}$ and $\mathbf{g}_i^{(\ell)}$, respectively. The block update follows a residual pre-norm design:
\begin{align}
  \mathbf{u}_i^{(\ell)} &= \mathrm{BN}\bigl(\mathbf{h}_i^{(\ell)} + \mathbf{g}_i^{(\ell)}\bigr),\\
  \tilde{\mathbf{u}}_i^{(\ell)} &= \mathrm{Dropout}\bigl(\mathrm{MLP}(\mathrm{LN}(\mathbf{u}_i^{(\ell)}))\bigr),\\
  \mathbf{x}_i^{(\ell+1)} &= \mathbf{x}_i^{(\ell)} + \tilde{\mathbf{u}}_i^{(\ell)}.
\end{align}
After $L_g$ blocks we set the node representation to $\mathbf{z}_i^{\mathrm{Graph}} = \mathbf{x}_i^{(L_g)}$.

\paragraph{Static Feature Encoder.}
Static patient attributes $\mathbf{x}_i^{\mathrm{Flat}}$ are embedded by a dedicated feedforward module comprising linear projection, LayerNorm, GELU activation, and dropout. The resulting vector $\mathbf{z}_i^{\mathrm{Flat}}$ defined in Equation (7) is mapped into the shared latent space and provides complementary context to dynamic and relational features.
\begin{equation}
  \mathbf{z}_i^{\mathrm{Flat}} = \mathrm{MLP}_{\mathrm{flat}}(\mathbf{x}_i^{\mathrm{Flat}})
\end{equation}

\paragraph{Fusion.}
To integrate information across temporal, relational, and static modalities, we perform a weighted concatenation of the corresponding embeddings. Let $\boldsymbol \lambda = (\lambda_{\mathrm{Graph}}, \lambda_{\mathrm{TS}}, \lambda_{\mathrm{Flat}})$ denote modality-specific importance weights, normalized via softmax. The fused representation is given by:
\begin{equation}
  \mathbf{z}_i^{\mathrm{fused}} = \mathrm{Concat}\left( \lambda_{\mathrm{Graph}}\,\mathbf{z}_i^{\mathrm{Graph}},\; \lambda_{\mathrm{TS}}\,\mathbf{z}_i^{\mathrm{TS}},\; \lambda_{\mathrm{Flat}}\,\mathbf{z}_i^{\mathrm{Flat}} \right).
\end{equation}

\paragraph{Prediction and Auxiliary Supervision.}
To reduce variance from outlier values and improve numerical stability, we \emph{supervise in the log domain}. Define the monotone transform
$T(y)=\log(1+y)$ and its inverse $T^{-1}(z)=\exp(z)-1$. Let $\tilde y_i=T(y_i)$.
The fused vector $\mathbf{z}_i^{\mathrm{fused}}$ is passed to a regression head that outputs a log-domain prediction $\tilde{\hat y}_i^{\mathrm{Main}}$; to encourage gradient flow, an auxiliary head on the time-series branch outputs $\tilde{\hat y}_i^{\mathrm{TS}}$. The training loss combines both outputs in the log domain:
\begin{equation}
    \mathcal{L}_i
    =(1-\alpha)\,\mathcal{L}_{\mathrm{Huber}}\!\big(\tilde{\hat y}_i^{\mathrm{Main}},\,\tilde y_i\big)
    +\alpha\,\mathcal{L}_{\mathrm{Huber}}\!\big(\tilde{\hat y}_i^{\mathrm{TS}},\,\tilde y_i\big),
\end{equation}
where $\alpha \in [0,1]$ controls the auxiliary weight. To mitigate the influence of extreme LOS values, a sample-dependent reweighting scheme is applied (based on the original-domain label):
\begin{equation}
    \mathcal{L}_i^{\mathrm{weighted}} = w(y_i)\,\mathcal{L}_i,\qquad
    w(y_i) = 1 + \gamma\,\mathbb{I}(y_i > \tau).
\end{equation}
During evaluation, we convert predictions back to the original time domain via
$\hat y_i = \max\{0,\,T^{-1}(\tilde{\hat y}_i^{\mathrm{Main}})\}$ and compute all metrics in that domain. Optimization is performed using AdamW with gradient clipping and early stopping based on validation $R^2$.

% \paragraph{Prediction and Auxiliary Supervision.}
% To reduce variance from outlier values and improve numerical stability, we apply a $\log(1 + y)$ transformation to all LOS labels during training. During evaluation, predicted values are converted back using $\exp(\cdot) - 1$ to compute metrics in the original time domain.
% The fused vector $\mathbf{z}_i^{\mathrm{fused}}$ is passed to a regression head to produce the main estimate $\hat{y}_i$. To encourage gradient flow and improve temporal feature learning, we attach an auxiliary head to the time-series branch, yielding an intermediate prediction $\hat{y}_i^{\mathrm{TS}}$. The training loss combines both outputs:
% \begin{equation}
%     \mathcal{L}_i = (1 - \alpha)\,\mathcal{L}_{\mathrm{Huber}}(\hat{y}_i^{\mathrm{Main}}, y_i)
%                    + \alpha\,\mathcal{L}_{\mathrm{Huber}}(\hat{y}_i^{\mathrm{TS}}, y_i)
% \end{equation}
% where $\alpha \in [0,1]$ controls the auxiliary weight. To mitigate the influence of extreme LOS values, a sample-dependent reweighting scheme is applied:
% \begin{equation}
%     \mathcal{L}_i^{\mathrm{weighted}} = w(y_i)\,\mathcal{L}_i,\qquad
%     w(y_i) = 1 + \gamma\,\mathbb{I}(y_i > \tau).
% \end{equation}
% Optimization is performed using AdamW with gradient clipping and early stopping based on validation $R^2$.

\section{Experimental Validation}

\paragraph{Data Description.}
We evaluate S$^2$G-Net on MIMIC-IV v3.1 dataset \cite{johnson2023mimic}, a large-scale collection of de-identified ICU records from Beth Israel Deaconess Medical Center. It comprises 65,347 adult patients. We extract 216 features, including 174 time-series and 42 static, aligned to hourly bins within a 48-hour window. Missing values are imputed via forward fill and decay masking. The dataset is split patient-wise into train (70\%), validation (15\%), and test (15\%).

\paragraph{Baselines.}
We compare our model with a baseline model suite covering three categories. (i) Sequential models include RNN~\cite{rumelhart1986learning}, BiLSTM~\cite{graves2005framewise}, and Transformer~\cite{vaswani2017attention} for temporal sequence modeling, along with Mamba~\cite{Gu23}, a recent state-space model optimized for efficient long-range dependency learning. (ii) Graph-based models operate on static patient similarity graphs, including Graph Attention Network (GAT)~\cite{velickovic2017graph}, Message Passing Neural Network (MPNN)~\cite{gilmer2017neural}, GraphSAGE~\cite{hamilton2017inductive}, and Optimized GraphGPS~\cite{rampavsek2022recipe}, which augments local message passing with global mixing via Mamba layers. (iii) Spatio-temporal models include LSTM-GNN~\cite{rocheteau2021predicting}, where patient time-series are encoded via LSTM and aggregated over static graphs, and dynamic LSTM-GNNs~\cite{rocheteau2021predicting}, which update graph structure during training to reflect evolving patient states. Results differ from Rocheteau et al. due to dataset (MIMIC-IV vs. eICU) and preprocessing differences but are used here as consistent baselines.
For completeness, XGBoost~\cite{chen2016xgboost} is included as a non-neural baseline for structured clinical prediction.

\paragraph{Evaluation Metrics.}
We report Mean Squared Error (MSE), Mean Squared Log Error (MSLE),  Mean Absolute Deviation (MAD), log-transformed Mean Absolute Percentage Error (log-MAPE), $R^2$ score, and Cohen’s linear weighted Kappa. MSE and $R^2$ are standard for regression; MSLE and log-MAPE are more robust to right-skewed LOS distributions. Kappa provides discrete agreement over stratified LOS bins. Formal definitions of all metrics are provided in Appendix~\ref{eval_metrics}.

\paragraph{Setups.}
We tune hyperparameters with Optuna~\cite{akiba2019optuna} over 75 trials, choosing the best configuration by validation $R^2$. The optimal setup adopts a 2-layer Mamba (128-dim) with a fusion coefficient of $\lambda=0.5$. The patient similarity graph is built via FAISS using $k_{\text{diag}}=3$, $k_{\text{BERT}}=1$, and MST.

\subsection{Results}

\begin{figure}[h]
\centering
\includegraphics[width=\textwidth]{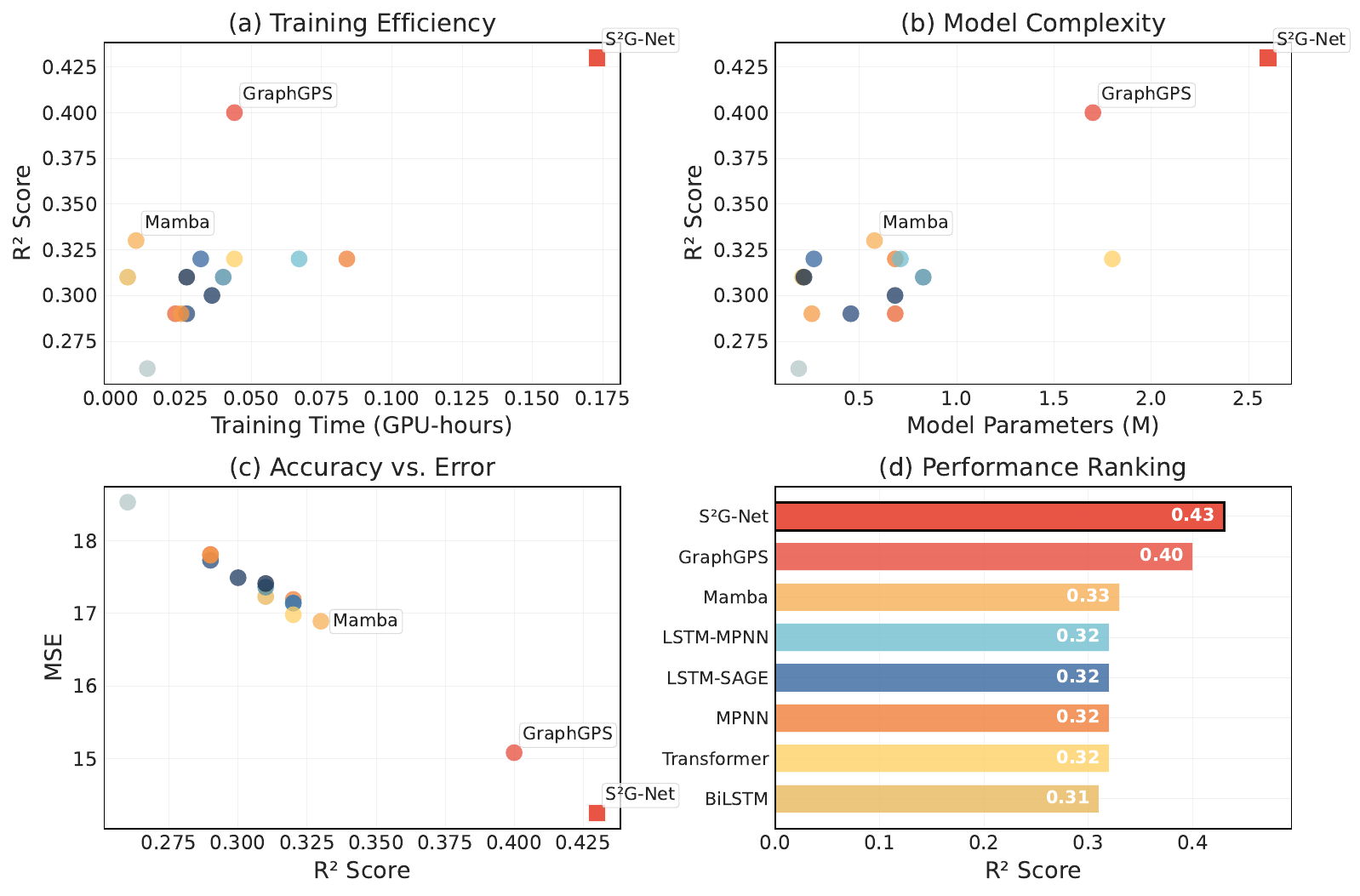}
\caption{Multi-dimensional analysis of S$^2$G-Net and baselines.
(a) Training cost vs. $R^2$; (b) Parameter count vs. $R^2$; (c) $R^2$ vs. MSE; (d) Overall ranking by $R^2$. XGBoost excluded for non-comparable computational metrics.}
\label{fig:performance_analysis}
\end{figure}

\paragraph{S$^2$G-Net Performance.}
As shown in Table~\ref{tab:performance_comparison} and Figure~\ref{fig:performance_analysis}, S$^2$G-Net consistently outperforms both neural and non-neural baselines across all 6 evaluation metrics. It obtains the highest $R^2$ ($0.43 \pm 0.01$) with statistically significant improvements in $R^2$, MSE, and Kappa. Compared to the best-performing baseline GraphGPS, S$^2$G-Net improves $R^2$ by around 7.5\% and reduces log-MAPE by 4.1\%. It also delivers competitive results on MSLE and MAD, indicating that integrating spatio-temporal graph model with domain-specific priors may offer benefits for clinical prediction tasks.

\begin{table*}[htbp]
\centering
\footnotesize
\setlength{\tabcolsep}{3.4pt}
\caption{Performance comparison of S$^2$G-Net and baseline models.
Mean $\!\pm\!$ standard deviation over 5 runs. \textsuperscript{†} indicates statistically significant improvement over the best baseline ($p < 0.05$).
Red highlights best results; orange indicates second-best. ↑ indicates higher is better; ↓ lower is better.}
\label{tab:performance_comparison}
\begin{tabular}{lcccccc}
\toprule
\textbf{Model}&\textbf{R\textsuperscript{2}}~$\uparrow$&\textbf{Kappa}~$\uparrow$&\textbf{MSE}~$\downarrow$&\textbf{MSLE}~$\downarrow$&\textbf{MAD}~$\downarrow$&\textbf{log-MAPE~(\%)}~$\downarrow$\\
\midrule
\multicolumn{7}{l}{\textit{Proposed Model}}\\
S$^2$G-Net & $\textcolor{red}{0.43\!\pm\!0.01}\textsuperscript{†}$ & $\textcolor{red}{0.42\!\pm\!0.00}\textsuperscript{†}$ & $\textcolor{red}{14.25\!\pm\!0.18}\textsuperscript{†}$ & $\textcolor{red}{0.25\!\pm\!0.01}$ & $\textcolor{red}{1.88\!\pm\!0.02}$ & $\textcolor{red}{35.74\!\pm\!1.24}$\\
\midrule
\multicolumn{7}{l}{\textit{Traditional Machine Learning Model}}\\
XGBoost & $0.32\!\pm\!0.00$ & $0.39\!\pm\!0.00$ & $17.32\!\pm\!0.03$ & $0.27\!\pm\!0.00$ & $1.96\!\pm\!0.01$ & $47.14\!\pm\!0.17$\\
\midrule
\multicolumn{7}{l}{\textit{Graph Neural Networks}}\\
GraphGPS& $\textcolor{orange}{0.40\!\pm\!0.01}$ & $\textcolor{orange}{0.40\!\pm\!0.00}$ & $\textcolor{orange}{15.08\!\pm\!0.24}$ & $\textcolor{orange}{0.27\!\pm\!0.02}$ & $\textcolor{orange}{1.93\!\pm\!0.05}$ & $\textcolor{orange}{37.27\!\pm\!1.26}$\\
GAT& $0.29\!\pm\!0.03$ & $0.34\!\pm\!0.02$ & $17.81\!\pm\!0.73$ & $0.28\!\pm\!0.01$ & $2.02\!\pm\!0.01$ & $43.98\!\pm\!3.76$\\
MPNN& $0.32\!\pm\!0.00$ & $0.36\!\pm\!0.00$ & $17.19\!\pm\!0.04$ & $0.28\!\pm\!0.00$ & $2.01\!\pm\!0.00$ & $41.20\!\pm\!0.83$\\
GraphSAGE& $0.29\!\pm\!0.03$ & $0.34\!\pm\!0.02$ & $17.80\!\pm\!0.64$ & $0.28\!\pm\!0.01$ & $2.03\!\pm\!0.01$ & $43.70\!\pm\!2.94$\\
\midrule
\multicolumn{7}{l}{\textit{Sequence Models}}\\
Mamba& $0.33\!\pm\!0.00$ & $0.36\!\pm\!0.00$ & $16.89\!\pm\!0.04$ & $0.28\!\pm\!0.02$ & $2.03\!\pm\!0.01$ & $40.46\!\pm\!1.03$\\
Transformer& $0.32\!\pm\!0.01$ & $0.36\!\pm\!0.00$ & $16.98\!\pm\!0.10$ & $0.29\!\pm\!0.01$ & $2.03\!\pm\!0.01$ & $44.80\!\pm\!1.20$\\
BiLSTM& $0.31\!\pm\!0.01$ & $0.35\!\pm\!0.01$ & $17.23\!\pm\!0.16$ & $0.28\!\pm\!0.00$ & $2.01\!\pm\!0.01$ & $42.09\!\pm\!2.38$\\
RNN& $0.26\!\pm\!0.02$ & $0.32\!\pm\!0.01$ & $18.53\!\pm\!0.45$ & $0.29\!\pm\!0.01$ & $2.02\!\pm\!0.01$ & $42.13\!\pm\!2.69$\\
\midrule
% \multicolumn{7}{l}{\textit{LSTM-GNN Models}}\\
\multicolumn{7}{l}{\textit{Hybrid Temporal-Graph Models}}\\
LSTM-MPNN& $0.32\!\pm\!0.03$ & $0.35\!\pm\!0.01$ & $17.14\!\pm\!0.68$ & $0.29\!\pm\!0.01$ & $2.02\!\pm\!0.02$ & $42.73\!\pm\!3.33$\\
LSTM-GAT& $0.31\!\pm\!0.01$ & $0.35\!\pm\!0.01$ & $17.36\!\pm\!0.21$ & $0.28\!\pm\!0.00$ & $2.01\!\pm\!0.00$ & $41.07\!\pm\!1.58$\\
LSTM-SAGE& $0.32\!\pm\!0.00$ & $0.36\!\pm\!0.00$ & $17.14\!\pm\!0.01$ & $0.29\!\pm\!0.00$ & $2.02\!\pm\!0.00$ & $42.81\!\pm\!1.98$\\
% \midrule
% \multicolumn{7}{l}{\textit{Dynamic LSTM-GNN Models}}\\
DyLSTM-GAT& $0.29\!\pm\!0.01$ & $0.33\!\pm\!0.01$ & $17.73\!\pm\!0.27$ & $0.28\!\pm\!0.01$ & $2.03\!\pm\!0.00$ & $42.83\!\pm\!1.30$\\
DyLSTM-MPNN& $0.30\!\pm\!0.01$ & $0.34\!\pm\!0.00$ & $17.49\!\pm\!0.29$ & $0.28\!\pm\!0.01$ & $2.02\!\pm\!0.00$ & $44.77\!\pm\!2.65$\\
DyLSTM-GCN& $0.31\!\pm\!0.01$ & $0.35\!\pm\!0.00$ & $17.41\!\pm\!0.21$ & $0.28\!\pm\!0.01$ & $2.03\!\pm\!0.02$ & $44.34\!\pm\!3.74$\\
\bottomrule
\end{tabular}
\end{table*}

\paragraph{Model Efficiency.}
Figure~\ref{fig:performance_analysis}a, b shows that S$^2$G-Net achieves a balanced trade-off between performance and computational cost. Compared to models of similar scale, it achieves higher $R^2$ in comparable or shorter training times without significant parameter overhead (less than 2.5M), making it a computationally efficient alternative for deployment in resource-constrained clinical settings.

\paragraph{Ablation Study.} Regarding modality contributions, excluding static features or using only static features each led to marked drops in model performance (Table~\ref{tab:ablation_results}). To obtain reliable predictions, both dynamic and static information sources are needed, and their integration is not trivially redundant.

\begin{figure*}[h]
    \centering
    \includegraphics[width=\linewidth]{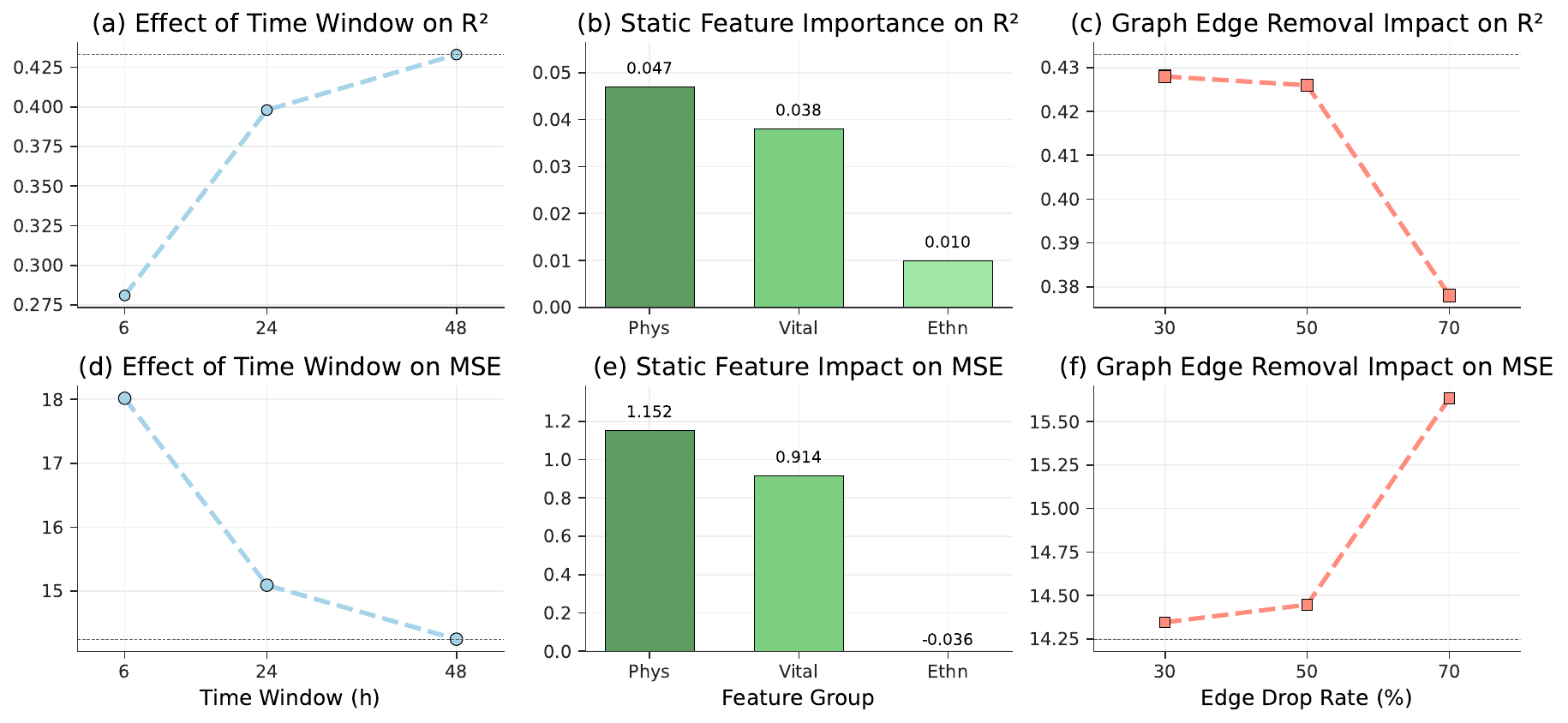}
    \caption{Ablation study on temporal, feature, and graph structure for $R^2$ and MSE.
    Dashed lines show the 48h baseline.
    (a,d) 6/24/48h input windows.
    (b,e) Removing static features Physiological (Phys), Vital signs (Vital), and Ethnicity (Ethn).
    (c,f) Edge dropout at 30\%, 50\%, and 70\%.
    }
    \label{fig:ablation-component-analysis}
\end{figure*}

Time window analysis reveals that longer observation windows provide substantial performance gains, especially between 6 and 24 hours (Figure~\ref{fig:ablation-component-analysis}a, d). Beyond 24 hours, the benefits plateau, possibly attributed to noise accumulation or reduced signal correlation in long-range dependencies.

\begin{figure*}[h]
    \centering
    \includegraphics[width=0.6\textwidth]{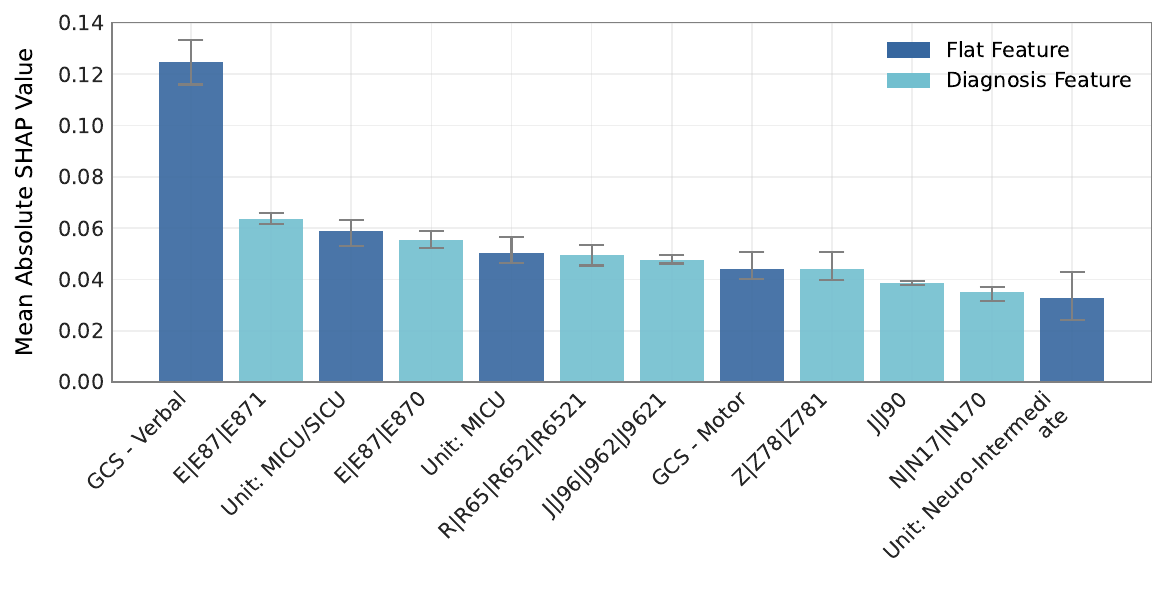}
    \caption{Top 12 features by SHAP importance, combining diagnosis-related and static features.}
    \label{fig:top15_all}
\end{figure*}

As illustrated in Figure~\ref{fig:ablation-component-analysis}b, e, Removal of physiological and vital sign features results in a notable performance drop. Among static features, ethnicity contributes the least to performance, while physiological and vital signals are more informative, as also verified by SHAP in Figure~\ref{fig:top15_all}.

\begin{table*}[ht]
\centering
\footnotesize
\setlength{\tabcolsep}{4.9pt}
\caption{Ablation study on temporal, feature, modality, and graph components.
Mean $\!\pm\!$ standard deviation over 3 random seeds.
↑ indicates higher is better; ↓ lower is better. Missing values denote negligible or undefined agreement. Baseline refers to full model with 48h input and all features.}
\label{tab:ablation_results}
\begin{tabular}{lcccccc}
\toprule
\textbf{Model}&\textbf{R\textsuperscript{2}}~$\uparrow$&\textbf{Kappa}~$\uparrow$&\textbf{MSE}~$\downarrow$&\textbf{MSLE}~$\downarrow$&\textbf{MAD}~$\downarrow$&\textbf{log-MAPE~(\%)}~$\downarrow$\\
\midrule
\multicolumn{7}{l}{\textit{Reference}} \\
Baseline & $0.43\!\pm\!0.00$ & $0.42\!\pm\!0.00$ & $14.25\!\pm\!0.00$ & $0.25\!\pm\!0.00$ & $1.88\!\pm\!0.00$ & $35.86\!\pm\!0.12$ \\
\midrule
\multicolumn{7}{l}{\textit{Temporal Window Analysis}} \\
Last 6h & $0.28\!\pm\!0.00$ & $0.33\!\pm\!0.00$ & $18.02\!\pm\!0.03$ & $0.39\!\pm\!0.00$ & $2.15\!\pm\!0.00$ & $45.09\!\pm\!0.21$ \\
Last 24h & $0.40\!\pm\!0.00$ & $0.40\!\pm\!0.00$ & $15.09\!\pm\!0.01$ & $0.28\!\pm\!0.00$ & $1.95\!\pm\!0.00$ & $37.77\!\pm\!0.13$ \\
\midrule
\multicolumn{7}{l}{\textit{Feature Group Impact}} \\
No Physiology & $0.39\!\pm\!0.00$ & $0.38\!\pm\!0.00$ & $15.40\!\pm\!0.00$ & $0.31\!\pm\!0.00$ & $2.12\!\pm\!0.00$ & $42.03\!\pm\!0.32$ \\
No Vitals & $0.40\!\pm\!0.00$ & $0.39\!\pm\!0.00$ & $15.16\!\pm\!0.00$ & $0.28\!\pm\!0.00$ & $2.00\!\pm\!0.00$ & $40.24\!\pm\!0.27$ \\
No Ethnicity & $0.42\!\pm\!0.00$ & $0.41\!\pm\!0.00$ & $14.21\!\pm\!0.01$ & $0.26\!\pm\!0.00$ & $1.91\!\pm\!0.00$ & $36.17\!\pm\!0.12$ \\
\midrule
\multicolumn{7}{l}{\textit{Modality Analysis}} \\
Static Only & $-$ & $-$ & $34.69\!\pm\!1.87$ & $1.32\!\pm\!0.44$ & $3.12\!\pm\!0.35$ & $93.66\!\pm\!10.73$ \\
No Static & $-$ & $-$ & $33.43\!\pm\!0.00$ & $0.99\!\pm\!0.00$ & $2.91\!\pm\!0.00$ & $105.54\!\pm\!17.01$ \\
\midrule
\multicolumn{7}{l}{\textit{Graph Robustness}} \\
Drop 30\% & $0.43\!\pm\!0.00$ & $0.41\!\pm\!0.00$ & $14.35\!\pm\!0.00$ & $0.25\!\pm\!0.00$ & $1.90\!\pm\!0.00$ & $36.90\!\pm\!0.11$ \\
Drop 50\% & $0.43\!\pm\!0.00$ & $0.41\!\pm\!0.00$ & $14.45\!\pm\!0.00$ & $0.27\!\pm\!0.00$ & $1.92\!\pm\!0.00$ & $37.23\!\pm\!0.16$ \\
Drop 70\% & $0.38\!\pm\!0.00$ & $0.37\!\pm\!0.00$ & $15.63\!\pm\!0.00$ & $0.29\!\pm\!0.00$ & $2.03\!\pm\!0.00$ & $42.67\!\pm\!0.31$ \\
\bottomrule
\end{tabular}
\end{table*}

Graph connectivity is further examined via controlled edge dropout (Figure~\ref{fig:ablation-component-analysis}c, f). Performance degrades gradually with increasing perturbation. The model retains a relatively stable $R^2$ up to 50\% edge removal. This indicate a degree of robustness to relational sparsity, a characteristic that is important when dealing with noisy or incomplete relationship data.

\paragraph{Interpretability.}
SHAP-based analysis revealed clinically meaningful predictors (Figure~\ref{fig:top15_all}). The high contribution of GCS language and motor scores is consistent with previous studies, confirming that neurological status is the ICU prognosis strong predictor~\cite{bastos1993glasgow}. ICD codes related to acute respiratory failure (J96.2) and acute kidney injury (N17.0) also match common ICU admission diagnoses and mortality risk factors~\cite{park2021acute, havaldar2024epidemiological}. The effect of ICU identifiers like MICU and SICU reflects differences in patient severity and treatment protocols by specialty. Its excellent calibration performance (ECE = 0.003) provides reliable predictive results for real-time clinical decision support.

\section{Conclusion}
In this work, we introduced S$^2$G-Net, a novel dual-path neural architecture for ICU LOS prediction. Our model uniquely combines SSMs to capture temporal patient data and GNNs to represent inter-patient relationships, leveraging this as a geometric prior. The results demonstrated that this hybrid approach consistently outperformed existing baselines on the large-scale MIMIC-IV dataset, proving to be a highly effective and computationally efficient solution for a critical task in healthcare. Importantly, this research validates the importance of a holistic approach that considers both an individual patient's trajectory and relationships within the patient population. Future work will include external validation of S$^2$G-Net on independent cohorts to assess generalizability and extending it to broader clinical tasks and real-time decision support to enhance patient care and resource efficiency.

\bibliographystyle{plainnat}
\bibliography{ref}

%%%%%%%%%%%%%%%%%%%%%%%%%%%%%%%%%%%%%%%%%%%%%%%%%%%%%%%%%%%%

\newpage
\appendix
\section{Compute Resources}
All experiments are conducted using Python 3.8 with CUDA 11.8. For reproducibility, training and evaluation require access to GPUs such as NVIDIA RTX 3090/4090 or Tesla V100/A100, each with $\geq$12 GB of VRAM. A minimum of 32 GB of system memory is necessary to process the full MIMIC-IV dataset, and approximately 60 GB of local storage is used to accommodate raw and processed files. Experiments are executed under Linux environments, with Windows supported through WSL2.

\section{Dataset Statistics}
\begin{table}[htbp]
\centering
\footnotesize
\caption{MIMIC-IV Cohort Statistics and Characteristics. 
Missing rate: 53.8\% for height; 
Other Units include Neuro ICU, Neuro Stepdown, and miscellaneous units. 
SD: Standard deviation. 
Ranges contain extreme outliers from raw data (e.g., weight up to 5,864 kg, height up to 445 cm).}
\label{tab:cohort_stats}
\begin{tabular}{p{2.2cm}p{1.7cm}p{1.2cm}|p{3.2cm}p{1.7cm}p{1.4cm}}
\toprule
\multicolumn{3}{c|}{\textbf{Dataset Overview}} & \multicolumn{3}{c}{\textbf{ICU Unit Distribution}} \\
\midrule
\textbf{Characteristic} & \textbf{Count} & \textbf{\%} & \textbf{ICU Unit} & \textbf{Count} & \textbf{\%} \\
\midrule
Total ICU Stays & 65,347 & 100.0 & Medical ICU & 12,732 & 19.5 \\
Training Set & 45,742 & 70.0 & Cardiac Vascular ICU & 11,520 & 17.6 \\
Validation Set & 9,802 & 15.0 & Med./Surgical ICU & 10,102 & 15.5 \\
Test Set & 9,803 & 15.0 & Surgical ICU & 8,980 & 13.7 \\
Temporal Records & 3,716,112 & $-$ & Trauma SICU & 7,791 & 11.9 \\
Static Features & 42 & $-$ & Coronary Care Unit & 7,127 & 10.9 \\
 &  &  & Other Units & 7,095 & 10.9 \\
\midrule
\multicolumn{3}{c|}{\textbf{ICU LOS Distribution}} & \multicolumn{3}{c}{\textbf{Patient Demographics}} \\
\midrule
\textbf{Statistic} & \textbf{Hours} & \textbf{Days} & \textbf{Characteristic} & \textbf{Value} & \textbf{Range/Dist.} \\
\midrule
Mean ± SD & 84.0 ± 122.4 & 3.5 ± 5.1 & Age (years), mean ± SD & 64.7 ± 17.1 & 18--103 \\
Median & 46.0 & 1.9 & Male Gender & 36,710 & 56.2\% \\
25th Percentile & 26.2 & 1.1 & Female Gender & 28,637 & 43.8\% \\
75th Percentile & 89.5 & 3.7 & Weight (kg), mean ± SD & 75.0 ± 39.4 & 1--5,864 \\
Range & 0.03--3,361.3 & 0.001--140.1 & Height (cm), mean ± SD & 169.3 ± 13.8 & 3--445 \\
\bottomrule
\end{tabular}%
\end{table}

\section{Evaluation Metrics}\label{eval_metrics}

\paragraph{Mean Squared Error (MSE).}MSE quantifies the average squared difference between the predicted and true LOS values, providing a standard measure of prediction accuracy:

\begin{equation*}
\text{MSE} = \frac{1}{N} \sum_{i=1}^{N} \left( y_i - \hat{y}_i \right)^2
\end{equation*}

where $y_i$ and $\hat{y}_i$ denote the true and predicted LOS for patient $i$.

\paragraph{Mean Squared Logarithmic Error (MSLE) \cite{pedregosa2011scikit}.}Given the right-skewed distribution of LOS in ICU cohorts, MSLE offers a more robust assessment by penalizing large relative errors while being less sensitive to extreme outliers:
\begin{equation*}
\text{MSLE} = \frac{1}{N} \sum_{i=1}^{N} \left( \log(y_i + 1) - \log(\hat{y}_i + 1) \right)^2
\end{equation*}

\paragraph{Mean Absolute Deviation (MAD) \cite{willmott2005advantages}.}MAD measures the average absolute error, providing an interpretable, outlier-robust assessment:
\begin{equation*}
\text{MAD} = \frac{1}{N} \sum_{i=1}^{N} | y_i - \hat{y}_i |
\end{equation*}

% \textbf{Mean Absolute Percentage Error (MAPE)} 
% MAPE evaluates the relative accuracy of predictions, defined as:
% \begin{equation*}
% \text{MAPE} = \frac{100\%}{N} \sum_{i=1}^{N} \left| \frac{y_i - \hat{y}_i}{y_i} \right|
% \end{equation*}
% To avoid division by zero, a minimum denominator threshold is applied for short stays.

\paragraph{Log-transformed Mean Absolute Percentage Error (log-MAPE).}
Define $\ell_i=\log(1+y_i)$ and $\hat{\ell}_i=\log(1+\hat{y}_i)$. Then
\begin{equation*}
\mathrm{MAPE}_{\log1p}
= \frac{100\%}{N}\sum_{i=1}^{N}
\frac{\left|\ell_i-\hat{\ell}_i\right|}{\max\{\ell_i,\epsilon\}} \, ,
\end{equation*}
where $\epsilon>0$ avoids division by zero.

\paragraph{Coefficient of Determination ($R^2$) \cite{pedregosa2011scikit}.}The $R^2$ score quantifies the proportion of variance in LOS explained by the model:
\begin{equation*}
R^2 = 1 - \frac{\sum_{i=1}^{N} (y_i - \hat{y}_i)^2}{\sum_{i=1}^{N} (y_i - \bar{y})^2}
\end{equation*}
where $\bar{y}$ is the mean observed LOS.

\paragraph{Linear Weighted Kappa (Kappa) \cite{cohen1968weighted}.}LOS predictions are discretized into 10 bins. The linear weighted kappa measures agreement between predicted and actual bins, accounting for chance agreement:
\begin{equation*}
\kappa_w = \frac{p_o^w - p_e^w}{1 - p_e^w}
\end{equation*}

where $p_o^w$ is the weighted observed agreement and $p_e^w$ is the weighted expected agreement by chance. Linear weights are assigned based on bin distance:

\begin{equation*}
w_{ij} = 1 - \frac{|i-j|}{k-1}
\end{equation*}

where $k=10$ bins and $i,j$ are bin indices. This weighting gives full credit for exact agreement, partial credit for near predictions, and penalizes distant disagreements proportionally.

\begin{table}[ht]
\centering
\footnotesize
\caption{Resource Consumption Comparison of Baseline Models. XGBoost is excluded as its computational metrics are not comparable to neural models.}
\label{tab:resource_comparison}
\begin{tabular}{lccc}
\toprule
\textbf{Model} & \textbf{Params} & \textbf{GPU-h} & \textbf{VRAM (GB)} \\
\midrule
\multicolumn{4}{l}{\textit{Proposed Method}}\\
S$^2$G-Net & 2.3M & 0.173 & 6.4 \\
\midrule
\multicolumn{4}{l}{\textit{Graph Neural Networks}}\\
GraphGPS & 1.7M & 0.044 & 5.3 \\
GAT & 820.7K & 0.023 & 2.6 \\
MPNN & 821.1K & 0.084 & 2.0 \\
GraphSAGE & 306.2K & 0.025 & 1.5 \\
\midrule
\multicolumn{4}{l}{\textit{Sequence Models}}\\
Mamba & 692.9K & 0.009 & 1.6 \\
Transformer & 1.8M & 0.060 & 2.4 \\
BiLSTM & 249.3K & 0.006 & 1.4 \\
RNN & 225.3K & 0.013 & 1.3 \\
\midrule
\multicolumn{4}{l}{\textit{Hybrid Temporal-Graph Models}}\\
LSTM-MPNN & 852.5K & 0.067 & 2.0 \\
LSTM-GAT & 993.4K & 0.040 & 3.2 \\
LSTM-SAGE & 318.6K & 0.032 & 2.9 \\
% \multicolumn{4}{l}{\textit{Dynamic LSTM-GNN Models}}\\
DyLSTM-GAT & 546.9K & 0.027 & 2.3 \\
DyLSTM-MPNN & 819.6K & 0.036 & 2.5 \\
DyLSTM-GCN & 257.7K & 0.027 & 1.8 \\
\bottomrule
\end{tabular}
\end{table}

\section{Efficiency and Resource Consumption}
Table~\ref{tab:resource_comparison} analyzes the memory and computational trade-offs of each approach. S$^2$G-Net integrates both a temporal state-space encoder and a relational graph module, keeping practical resource consumption: it requires 2.3M parameters, 0.173 GPU-hours per run, and a peak VRAM of 6.4 GB. This is modestly higher than GraphGPS (1.7M params, 5.3 GB VRAM) and substantially lower than ensemble or multi-stage architectures that are often infeasible in real-world deployments.

\section{Hyperparameter Sensitivity}
\begin{figure*}[ht]
\centering
\includegraphics[width=\textwidth]{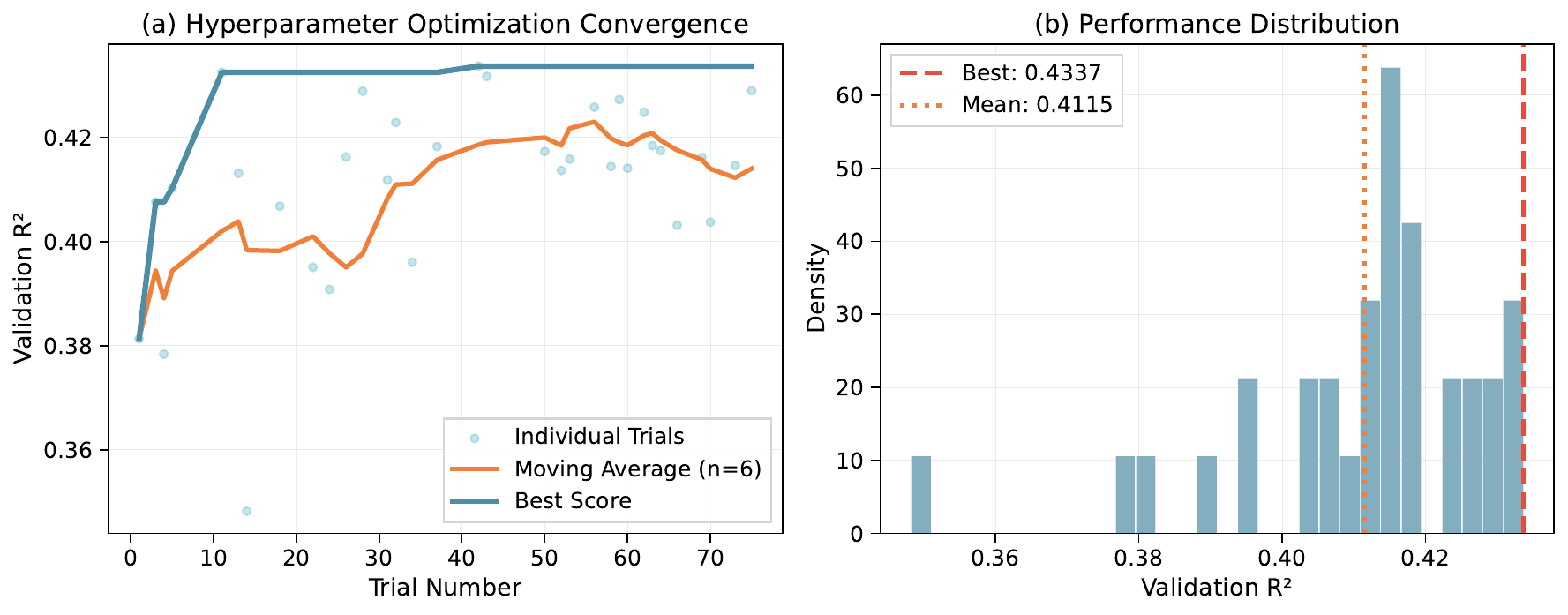}
\caption{
Hyperparameter optimization convergence.
(a) Validation $R^2$ scores over 75 trials (excluding failures with $R^2 \leq 0$), including individual scores, an 8-trial moving average, and the best score.
(b) Distribution of validation $R^2$ scores. The red dashed line marks the best score ($0.4337$), and the dotted line shows the mean ($0.4115$).
}
\label{fig:optimization_convergence}
\end{figure*}

We performed 75 independent trials using Optuna's tree-structured Parzen estimator (TPE)~\cite{akiba2019optuna}. Figure~\ref{fig:optimization_convergence}a shows that validation $R^2$ increases steadily across 75 optimization trials, improving rapidly at first and tapering after approximately 12 trials. The performance histogram (Figure~\ref{fig:optimization_convergence}b) demonstrates a unimodal distribution, suggesting stable search behavior and the absence of degenerate hyperparameter regions.

\begin{figure}[h]
\centering
\includegraphics[width=0.75\textwidth]{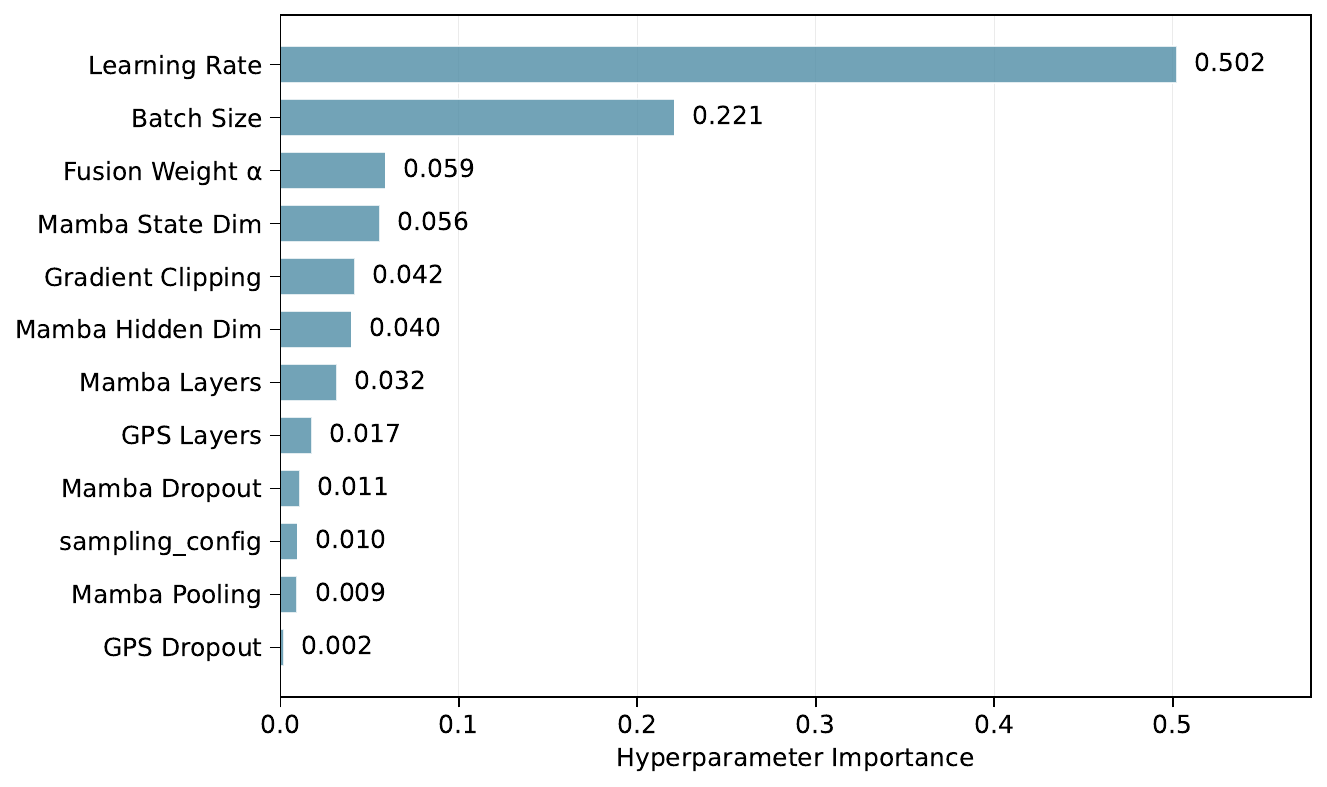}
\caption{Hyperparameter sensitivity.
Importance scores reflect each hyperparameter's impact on validation $R^2$ using Optuna's TPE.}
\label{fig:hyperparam_sensitivity}
\end{figure}

Figure \ref{fig:hyperparam_sensitivity} shows hyperparameter importance. The learning rate dominates (0.502), indicating the performance of S$^2$G-Net is highly sensitive to this setting, with suboptimal values causing marked degradation. Batch size (0.221) is the second most influential but far less critical. All other parameters, including fusion ($\lambda$), model depth, hidden/state dimensions, and regularization or pooling, each contribute under 6\% to validation $R^2$ variance and can be tuned within broad ranges without substantial impact.

\section{Hyperparameter Tuning}
Table \ref{tab:hparam_space} shows that the optimal S$^2$G-Net model used a 2-layer Mamba encoder with 128 hidden dimensions and 16 state dimensions, a 2-layer GraphGPS module, dropout rates of 0.1, and last pooling in the time-series encoder. The fusion coefficient $\lambda$ was set to 0.5, the learning rate to $1.7\times 10^{-4}$, and the batch size to 32. The neighborhood sampling sizes are 15 and 10 for the first and second hops, respectively. The learning rate was sampled log-uniformly in $[1\times 10^{-5},\,1\times 10^{-3}]$ using Optuna's TPE; the best value selected by validation RMSE ($\mathrm{RMSE}=\sqrt{\mathrm{MSE}}$) was $1.7\times 10^{-4}$. Hyperparameter optimization is conducted strictly on the validation set; the test set was used only for final model evaluation. 

\begin{table}[htbp]
    \centering
    \caption{Hyperparameter search space and the best-performing values selected by validation RMSE for S$^2$G-Net.}
    \label{tab:hparam_space}
    \begin{tabular}{lp{3.5cm}l}
        \toprule
        \textbf{Parameter} & \textbf{Search Space} & \textbf{Best Value} \\
        \midrule
        Mamba d\_model              & 64, 128, 256              & 128 \\
        Mamba layers                & 2, 3, 4                   & 2 \\
        Mamba d\_state              & 16, 32, 64                & 16 \\
        Mamba dropout               & 0.1, 0.2                  & 0.1 \\
        Mamba pooling               & mean, last                & last \\
        GPS layers                  & 2, 3, 4                   & 2 \\
        GPS dropout                 & 0.1, 0.2                  & 0.1 \\
        Fusion $\lambda$            & 0.3, 0.5, 0.7             & 0.5 \\
        Learning rate               & log-uniform $[1\times10^{-5},\,1\times10^{-3}]$ & $1.7\times10^{-4}$ \\
        Batch size                  & 32, 64, 128             & 32 \\
        Gradient clipping           & 0.0, 2.0, 5.0             & 2.0 \\
        Sampling config (N, L)      & (15, 10), (25, 15)        & (15, 10) \\
        \bottomrule
    \end{tabular}
\end{table}

\section{Graph Construction Analysis}
Table~\ref{tab:top_graph_configs} shows 10 highest-scoring graph construction. The FAISS and TF-IDF methods are heavily weighted in the top results, while the MST reconnection strategy appears in high-ranking configurations. It indicates that S$^2$G-Net benefits from multi-view edges and judicious use of sparse connectivity, rather than indiscriminately increasing graph density. Moreover, small changes in $k$ or similarity methods typically result in only minor differences in performance.

\begin{table}[ht]
\centering
\caption{Top 10 graph-construction settings ranked by $R^{2}$. Density is scaled by $10^{-6}$ for readability; $|E|$ denotes edge count.}
\label{tab:top_graph_configs}
\setlength{\tabcolsep}{5pt}
\begin{tabular}{cccc|cc|c}
\toprule
\boldmath$k_{\text{diag}}$ & \boldmath$k_{\text{BERT}}$ & \textbf{Edge Method} & \textbf{Rewiring} & \boldmath$|E|$ & \textbf{Density} & \boldmath$R^{2}$ $\uparrow$ \\
\midrule
3 & 1 & FAISS & MST & 206,847 & 96.88 & 0.4375 \\
1 & 3 & Penalize & None & 237,179 & 111.09 & 0.4361 \\
1 & 1 & TF-IDF & MST & 107,057 & 50.14 & 0.4321 \\
1 & 3 & FAISS & None & 242,833 & 113.73 & 0.4304 \\
1 & 1 & FAISS & None & 112,231 & 52.57 & 0.4293 \\
5 & 1 & FAISS & MST & 294,081 & 137.74 & 0.4272 \\
1 & 1 & TF-IDF & None & 107,057 & 50.14 & 0.4271 \\
3 & 1 & TF-IDF & None & 186,078 & 87.15 & 0.4264 \\
1 & 1 & Penalize & None & 106,614 & 49.93 & 0.4249 \\
3 & 3 & TF-IDF & GDC & 316,726 & 148.34 & 0.4241 \\
\bottomrule
\end{tabular}
\end{table}

\section{Reliability and Calibration}
\begin{figure}[ht]
    \centering
    \includegraphics[width=0.75\textwidth]{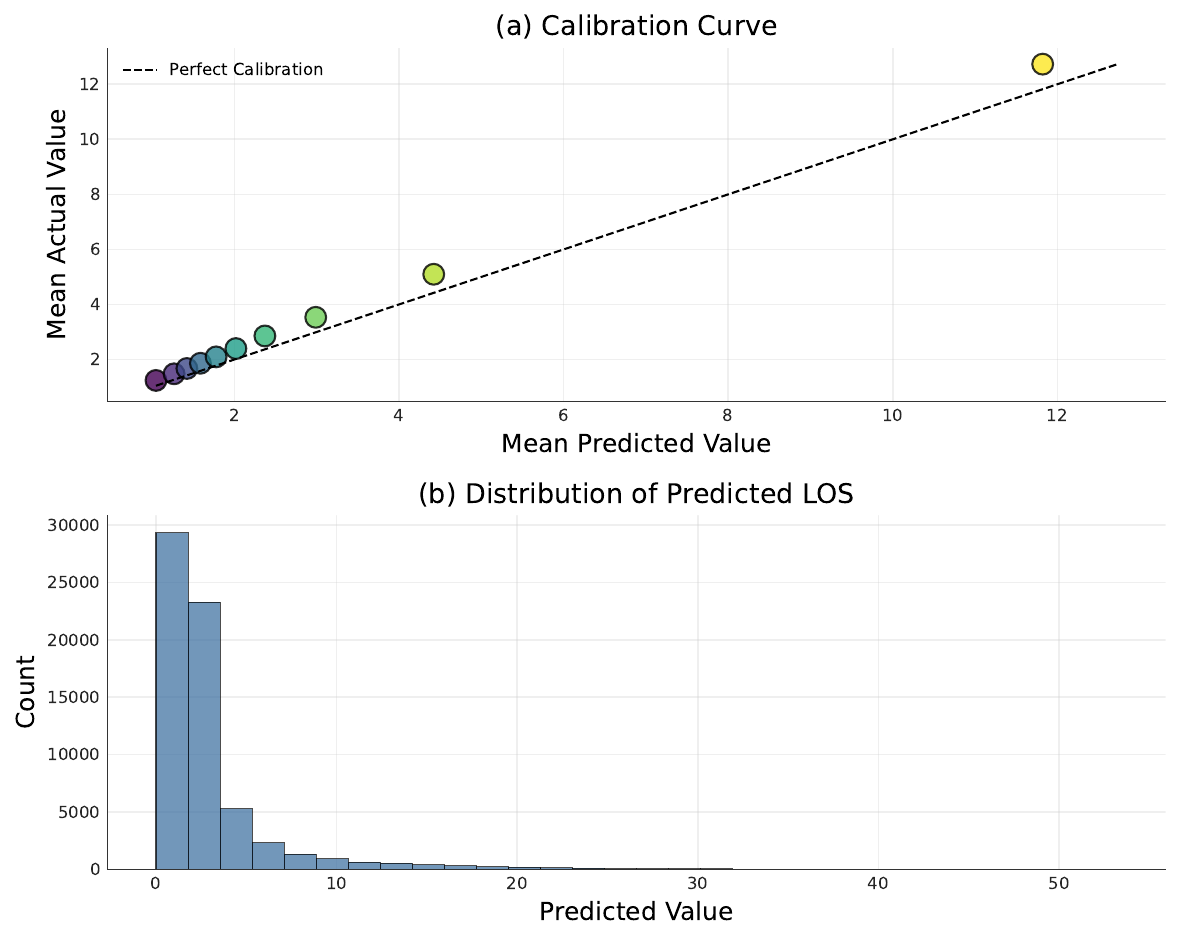}
    \caption{Reliability diagram comparing predicted and actual values across prediction bins, with a distribution histogram.}
    \label{fig:reliability}
\end{figure}

We evaluated the probabilistic predictions of S$^2$G-Net using a calibration plot (Figure~\ref{fig:reliability}). The curve highly matches the actual LOS values within each interval, giving a standardized expected calibration error of 0.003, showing that the calibration error is minimal. The accompanying histogram further confirms the good calibration of the prediction results within the clinically relevant LOS range.

\end{document}